%% file: main.tex
\documentclass{article}
\usepackage{balance}
\usepackage{url}
\usepackage{bm}
\usepackage[colorlinks=true,citecolor=blue,anchorcolor=purple]{hyperref}
\usepackage{spconf,amsmath,graphicx, booktabs}

\renewenvironment{thebibliography}[1]{
  \begin{oldthebibliography}{#1}
    \setlength{\itemsep}{0.3ex}
    \setlength{\parskip}{0.3ex}
}
{
  \end{oldthebibliography}
}
\usepackage{wrapfig}
\usepackage{booktabs}
\usepackage{chemformula}
\usepackage{amsmath}
\usepackage{amssymb}
\usepackage{tcolorbox}
\usepackage{color,soul}
\usepackage{algorithm}
\usepackage{algpseudocode}

\newtcbox{\entoure}[1][red]{on line,
arc=3pt,colback=#1!10!white,colframe=#1!50!black,
before upper={\rule[-3pt]{0pt}{10pt}},boxrule=1pt,
boxsep=0pt,left=2pt,right=2pt,top=1pt,bottom=.5pt}
\newcommand{\ecnas}{{\tt EC-NAS}}

\def\vec#1{\mathchoice{\mbox{\boldmath$\displaystyle#1$}}
{\mbox{\boldmath$\textstyle#1$}}
{\mbox{\boldmath$\scriptstyle#1$}}
{\mbox{\boldmath$\scriptscriptstyle#1$}}}

\newcommand{\ndom}{\operatorname{ndom}}
\newcommand{\HYP}{\operatorname{\mathcal{S}}}
\newcommand{\VOL}{\operatorname{\Lambda}}
\newcommand{\CON}{\operatorname{\Delta}}
\newcommand{\B}{X''}
\newcommand{\A}{X'}
\newcommand{\F}{\mathcal{F}}
\newcommand{\X}{\mathbb{X}}

\title{EC-NAS: Energy Consumption Aware Tabular Benchmarks for Neural Architecture Search}
\name{Pedram Bakhtiarifard$^{\dagger}$ \qquad Christian Igel$^{\dagger}$ \qquad Raghavendra Selvan$^{\dagger}$\sthanks{The authors acknowledge funding received under European Union’s Horizon Europe Research and Innovation programme under grant agreements No. 101070284 and No. 101070408.}}

\address{$^{\dagger}$  Department of Computer Science, University of Copenhagen, Denmark}
\begin{document}
%\ninept
%
\maketitle
%
% TODO: Mention something about the algorithm devised
\begin{abstract}
Energy consumption from the selection, training, and deployment of deep learning models has seen a significant uptick recently. This work aims to facilitate the design of energy-efficient deep learning models that require less computational resources and prioritize environmental sustainability by focusing on the energy consumption. Neural architecture search (NAS) benefits from tabular benchmarks, which evaluate NAS strategies cost-effectively through pre-computed performance statistics. We advocate for including energy efficiency as an additional performance criterion in NAS. To this end, we introduce an enhanced tabular benchmark encompassing data on energy consumption for varied architectures. The benchmark, designated as \ecnas{}\footnote{Source code is available at: \url{https://github.com/saintslab/EC-NAS-Bench}}, has been made available in an open-source format to advance research in energy-conscious NAS.
\ecnas{} incorporates a surrogate model to predict energy consumption, aiding in diminishing the energy expenditure of the dataset creation. Our findings emphasize the potential of \ecnas{} by leveraging multi-objective optimization algorithms, revealing a balance between energy usage and accuracy. This suggests the feasibility of identifying energy-lean architectures with little or no compromise in performance.
\end{abstract}

\vspace{-0.5cm}
\section{Introduction}
\label{sec:intro}
    \vspace{-0.25cm}
\begin{figure}[htb]
    \centering
    \includegraphics[width=0.5\textwidth]{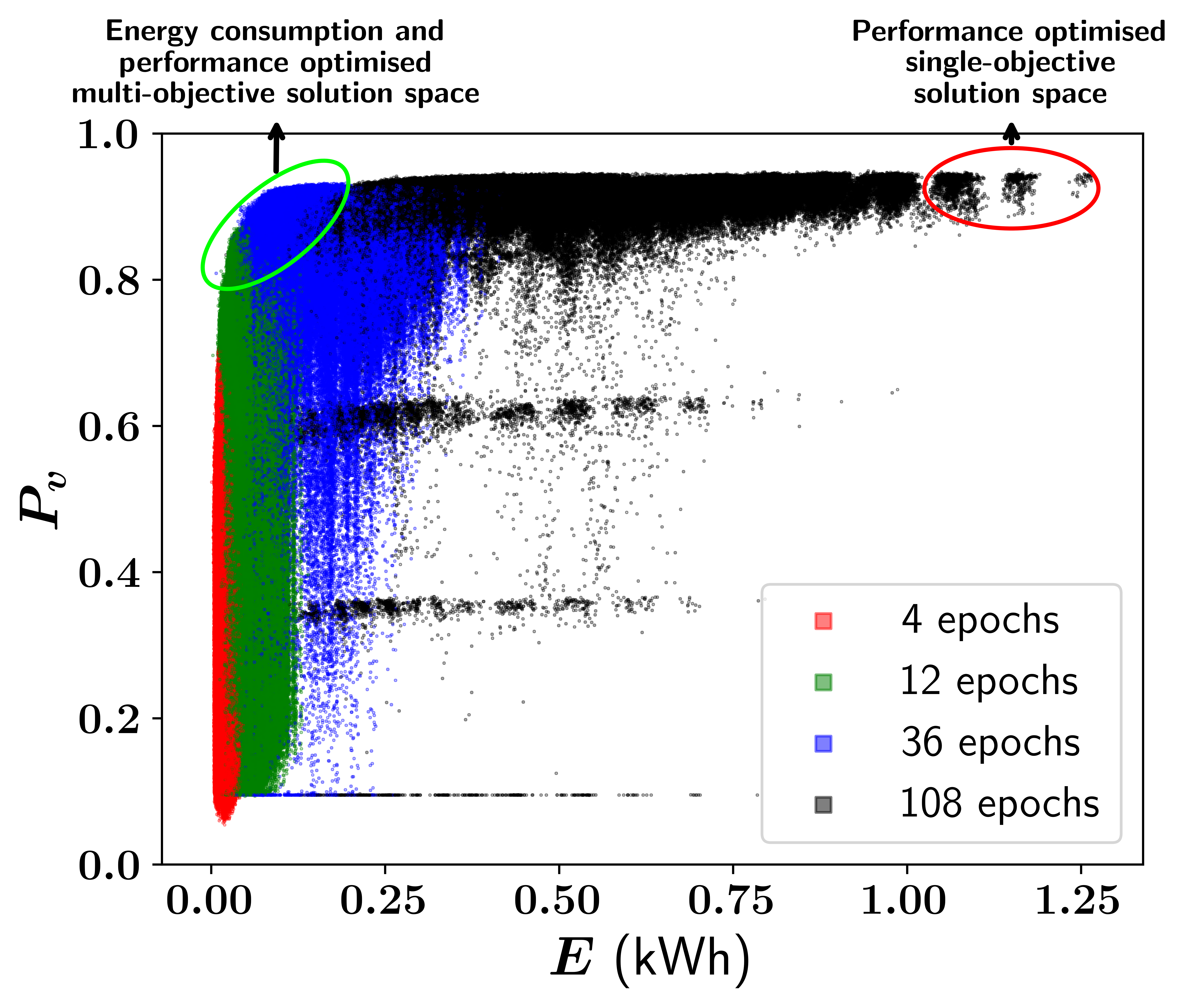}
        \vspace{-1cm}
    \caption{Scatter plot of about 423k CNN architectures showing training energy ($E$) vs. validation performance ($P_v$) across four training budgets. Solutions in the top-right (red ellipse) prioritize performance at high energy costs. Joint optimization shifts preferred solutions to the left (green ellipse), indicating reduced energy with minimal performance loss.}
    \label{fig:intro}
    \vspace{-0.75cm}
\end{figure}

Neural Architecture Search (NAS) strategies, which explore model architectures based on training and evaluation metrics, have demonstrated their ability to reveal novel designs with state-of-the-art performance \cite{ren2020comprehensive, sota, nas_pred1}. While promising, NAS comes with computational and energy-intensive demands, leading to significant environmental concerns due to the carbon footprint incurred due to energy consumption~\cite{tan2019efficientnet, greenAI, anthony2020carbontracker}. Given the rapidly increasing computational requirements of deep learning models \cite{sevilla2022compute}, there is an imperative to address the balance between performance and resource efficiency.

Efficient evaluation of NAS strategies has gained traction, using pre-computed performance statistics in tabular benchmarks and the use of surrogate and one-shot models \cite{nas_benchmarks, zela2022surrogate, nasBench101Dataset, sota, nasbench1shot1}. 
Nevertheless, the primary focus remains on performance, with the trade-offs between performance and energy efficiency often overlooked. This trade-off is visually represented in Figure \ref{fig:intro}, illustrating the potential to find energy-efficient models without compromising performance. Aligning with recent advancements in energy-aware NAS research, we advocate for integrating energy consumption as a pivotal metric in tabular NAS benchmarks. We aim to uncover inherently efficient deep learning models, leveraging pre-computed energy statistics for sustainable model discovery. This perspective is supported by recent works, such as the EA-HAS-Bench \cite{dou2023eahasbench}, which emphasizes the trade-offs between performance and energy consumption. Furthermore, the diverse applications of NAS in areas like speech emotion recognition \cite{wu2022neural} and visual-inertial odometry \cite{10095166} underscore its versatility and the need for efficiency.

%\vspace{-0.25cm}
\section{Energy Awareness in NAS}
\vspace{-0.25cm}

Building upon the foundational NAS-Bench-101 \cite{nasBench101Dataset}, we introduce our benchmark, \ecnas, to accentuate the imperative of energy efficiency in NAS. Our adaptation of this dataset, initially computed using an exorbitant 100 TPU years equivalent of compute time, serves our broader mission of steering NAS methodologies towards energy consumption awareness.

\vspace{-0.25cm}
\subsection{Architectural Design and Blueprint} \label{sec:arch}
Central to our method are architectures tailored for CIFAR-10 image classification \cite{cifar10}. We introduce additional objectives for emphasizing the significance of hardware-specific efficiency trends in deep learning models. The architectural space is confined to the topological space of cells, with each cell being a configurable feedforward network. In terms of cell encoding, these individual cells are represented as directed acyclic graphs (DAGs). Each DAG, $G(V,M)$, has $N=|V|$ vertices (or nodes) and edges described in a binary adjacency matrix $M\in \{0,1\}^{N\times N}$. The set of operations (labels) that each node can realise is given by $\mathcal{L}^{\prime}=\{{\tt input,output}\} \cup \mathcal{L}$, where $\mathcal{L} = \{{\tt 3x3conv,1x1conv,3x3maxpool}\}$. Two of the $N$ nodes are always fixed as ${\tt input}$ and ${\tt output}$ to the network. The remaining $N-2$ nodes can take up one of the labels in $\mathcal{L}$. The connections between nodes of the DAG are encoded in the upper-triangular adjacency matrix with no self-connections (zero main diagonal entries). For a given architecture, $\mathcal{A}$, every entry $\alpha_{i,j} \in M_{\mathcal{A}}$ denotes an edge, from node $i$ to node $j$ with operations $i,j \in \mathcal{L}$ and its labelled adjacency matrix, $L_\mathcal{A} \in M_{\mathcal{A}} \times \mathcal{L}^\prime$.

\vspace{-0.25cm}
\subsection{Energy Measures in NAS}
\label{sec:energy_measures}

Traditional benchmarks, while insightful, often fall short of providing a complete energy consumption profile. In \ecnas, we bring the significance of energy meaures to the forefront, crafting a comprehensive view that synthesizes both hardware and software intricacies. The mainstays of neural network training -- GPUs and TPUs -- are notorious for their high energy consumption \cite{anthony2020carbontracker,dodge2022measuring}. To capture these nuances, we utilize and adopt the {\tt Carbontracker} tool \cite{anthony2020carbontracker} to our specific needs, allowing us to observe total energy costs, computational times, and aggregate carbon footprints.

\vspace{-0.25cm}
\subsection{Surrogate Model for Energy Estimation} \label{sec:surrogate}

The landscape of NAS has transformed to encompass a broader spectrum of metrics. Energy consumption, pivotal during model training, offers insights beyond the purview of traditional measures such as floating-point operations (FPOPs) and computational time. Given the variability in computational time, owing to diverse factors like parallel infrastructure, this metric can occasionally be misleading. Energy consumption, in contrast, lends itself as a more consistent and comprehensive measure, factoring in software and hardware variations. We measure the energy consumption of training the architectures on the CIFAR-10 dataset, following the protocols to NAS-Bench-101. The in-house SLURM cluster, powered by an NVIDIA Quadro RTX 6000 GPU and two Intel CPUs, provides an optimal environment.

\begin{figure}[t]
    \centering
    \includegraphics[width=0.239\textwidth]{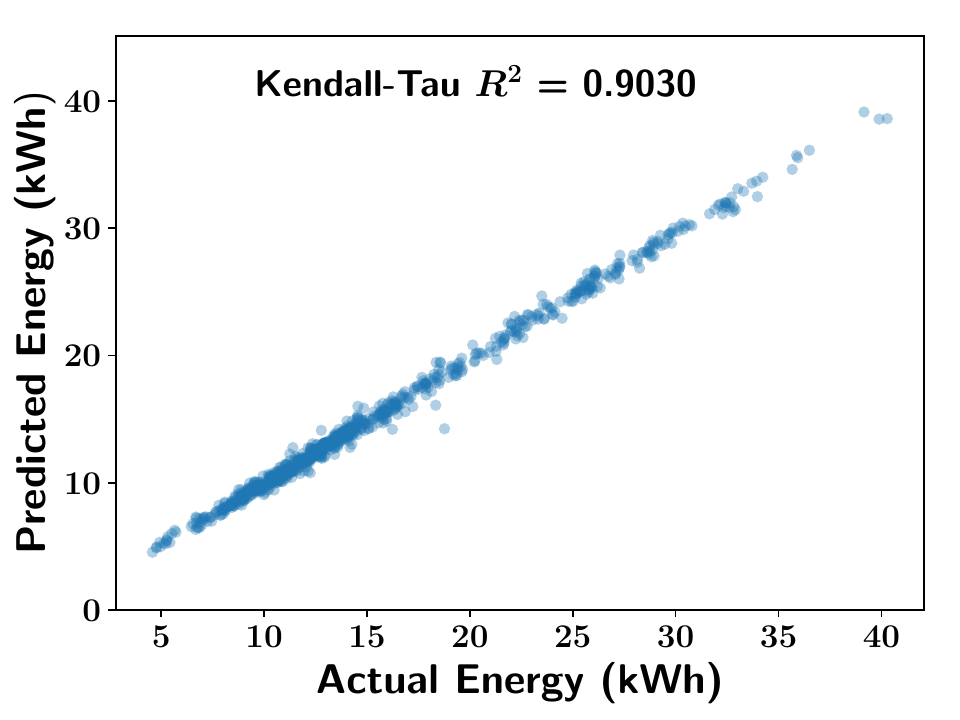}
    \includegraphics[width=0.239\textwidth]{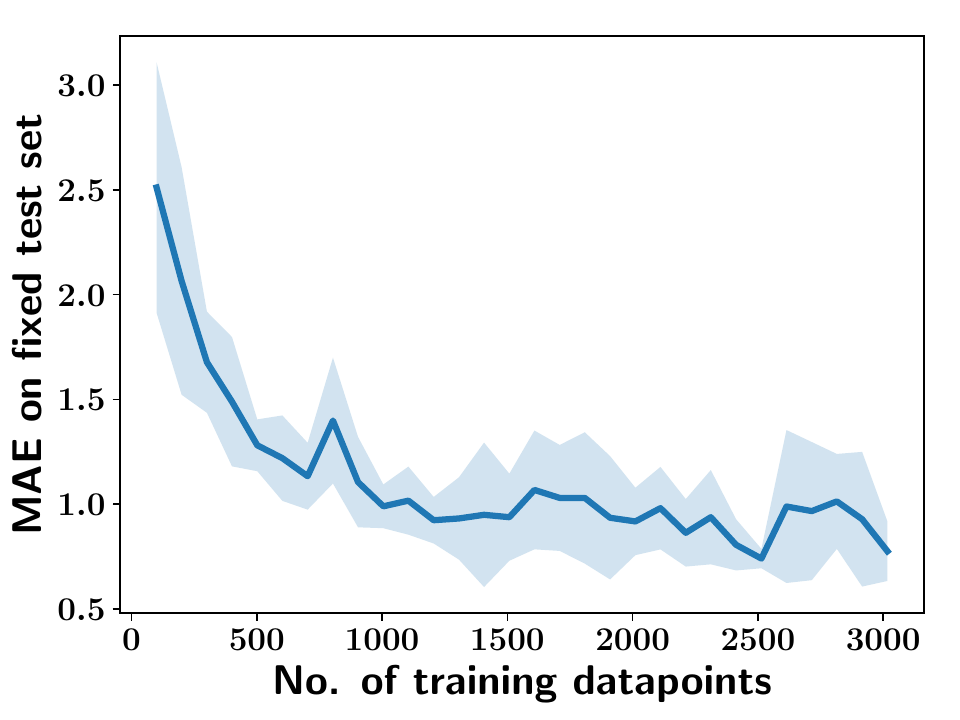}
    \vspace{-0.75cm}
    \caption{Scatter plot depicting the Kendall-Tau correlation coefficient between predicted and actual energy consumption (left) and the influence of training data size on test accuracy (right). Error bars are based on 10 random initializations.}
    \label{fig:surrogate}
    \vspace{-0.25cm}
\end{figure}

The vast architecture space, however, introduces challenges in the direct energy estimation. Our remedy to this is a surrogate model approach, wherein we derived insights to guide a multi-layer perceptron (MLP) model by training using a representative subset of architectures. This surrogate model adeptly predicts energy consumption patterns, bridging computational demand and energy efficiency. Its efficacy is highlighted by the strong correlation between its predictions and actual energy consumption values, as illustrated in Figure~\ref{fig:surrogate}.

\begin{figure*}[t]
    \centering
    \includegraphics[width=0.245\textwidth]{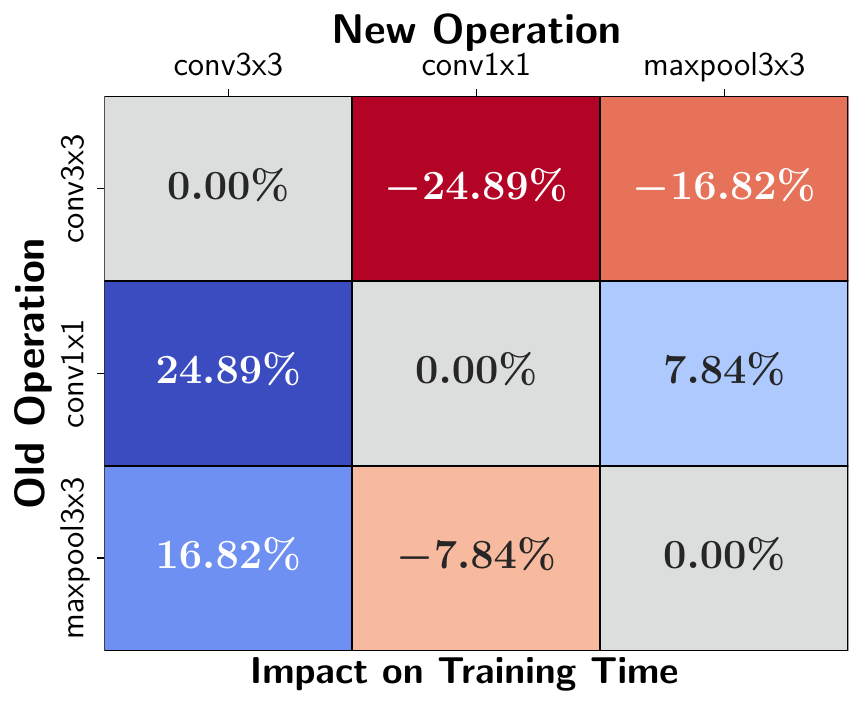}
    \includegraphics[width=0.245\textwidth]{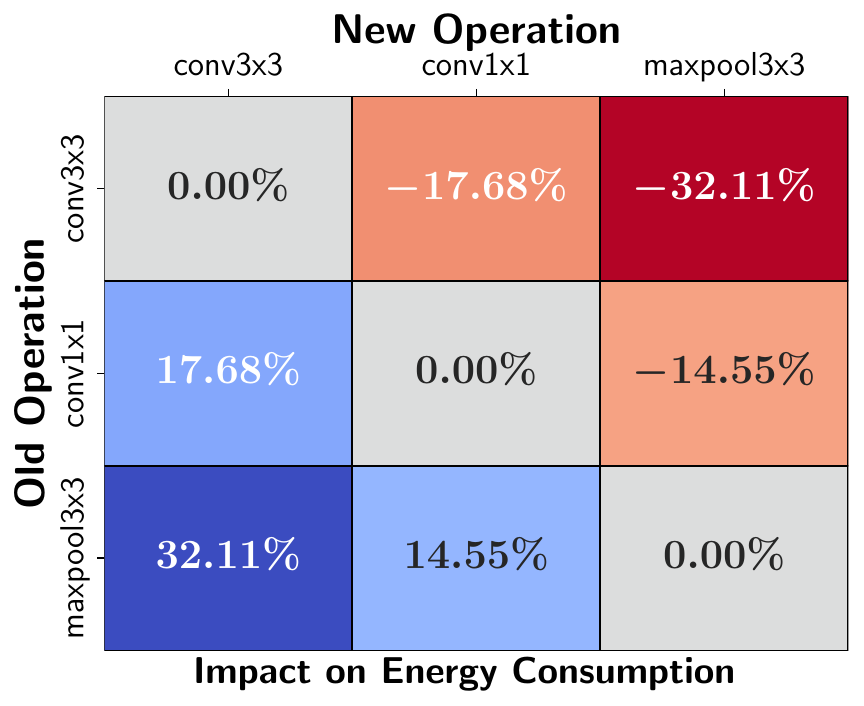}
    \includegraphics[width=0.245\textwidth]{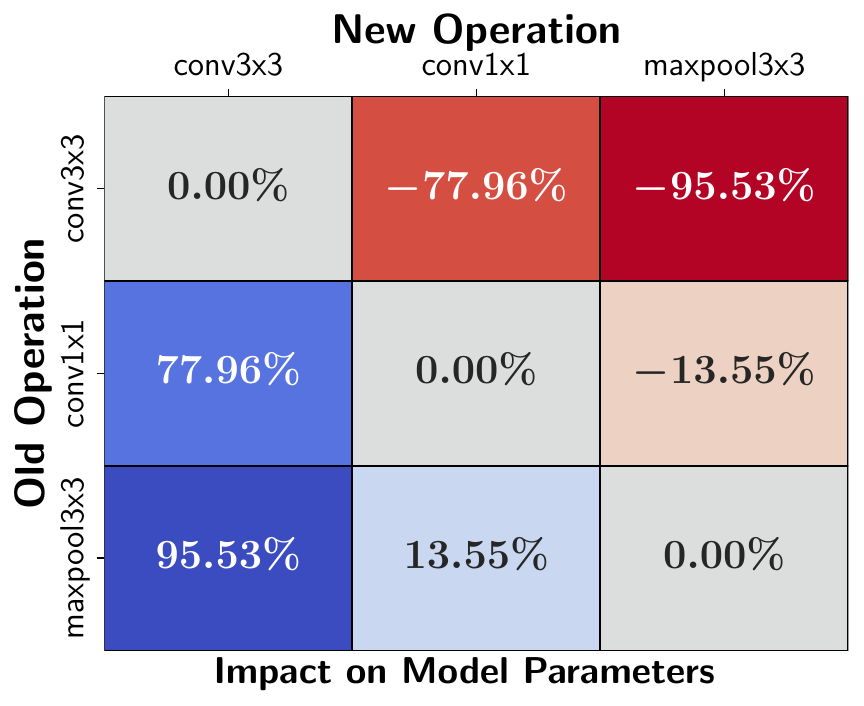}
    \includegraphics[width=0.245\textwidth]{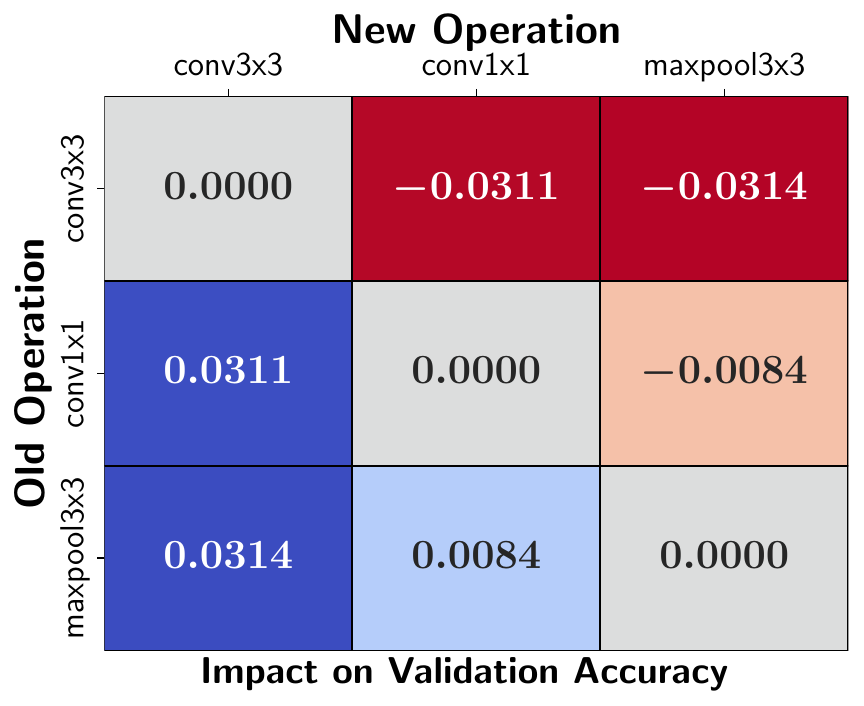}
    \vspace{-0.5cm}
    \caption{Aggregated impact of swapping one operator for another on energy consumption, training time, validation accuracy, and parameter count. The figure illustrates how changing a single operator can affect the different aspects of model performance, emphasizing the importance of selecting the appropriate operators to balance energy efficiency and performance.}
    \label{fig:aggr}
    \vspace{-0.25cm}
\end{figure*}

\vspace{-0.25cm}
\subsection{Dataset Analysis and Hardware Consistency}
Understanding architectural characteristics and the trade-offs they introduce is crucial. This involves studying operations, their impacts on efficiency and performance, as well as the overarching influence of hardware on energy costs. Training time and energy consumption trends naturally increase with model size. However, gains in performance tend to plateau for models characterized by larger DAGs. Interestingly, while parameter variation across model sizes remains minimal, training time and energy consumption show more significant variability for more extensive models. These findings highlight the multifaceted factors affecting performance and efficiency.

Different operations can also have a profound impact on performance. For instance, specific operation replacements significantly boost validation accuracy while increasing energy consumption without increasing training time. This complex relationship between training time, energy consumption and performance underscore the importance of a comprehensive approach in NAS. The impact of swapping one operation for another on various metrics, including energy consumption, training time, validation accuracy, and parameter count, is captured in Figure~\ref{fig:aggr}.

\begin{figure}[t]
\centering
\includegraphics[width=0.45\textwidth]{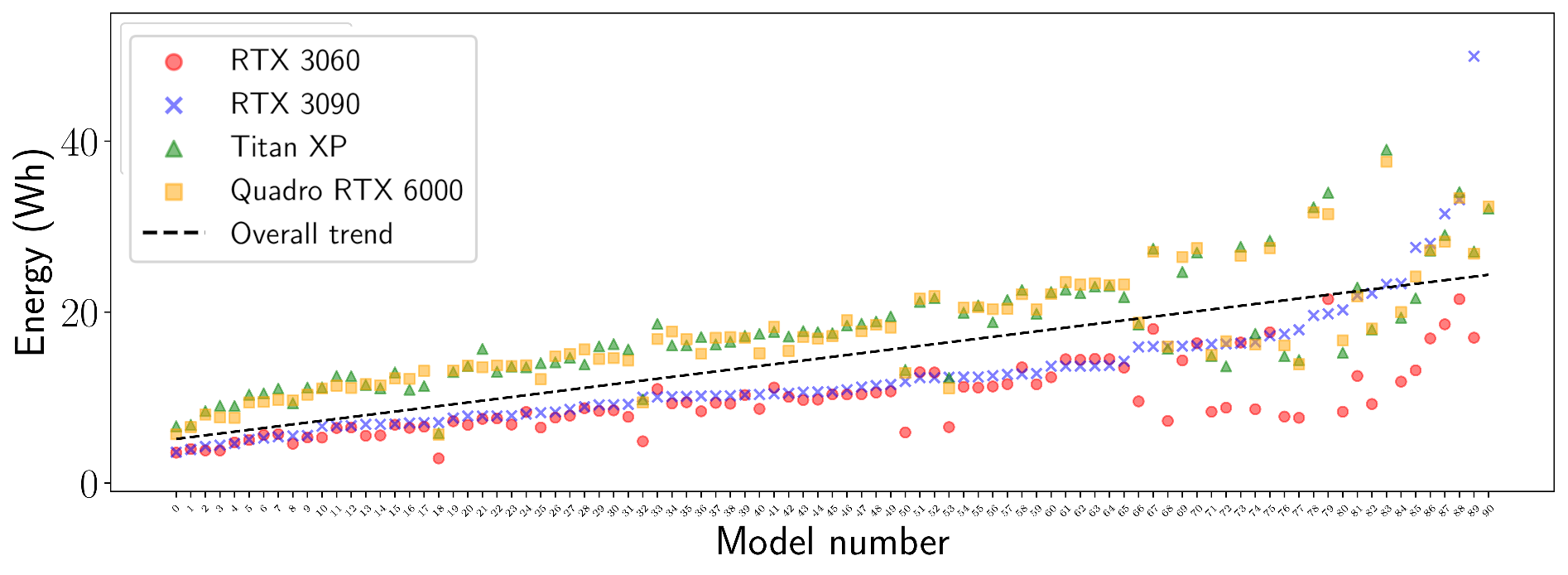}
\vspace{-0.5cm}
    \caption{Energy consumption of models with DAGs where \( |V| \leq 4 \) on different GPUs. Models are organized by their average energy consumption for clarity.}
\label{fig:hardware}
\vspace{-0.5cm}
\end{figure}

In \ecnas, we further probed the energy consumption patterns of models characterized by DAGs with $|V| \le 4$, spanning various GPUs. This exploration, depicted in Figure~\ref{fig:hardware}, confirms the flexibility of the benchmark across different hardware environments. This adaptability paves the way for advanced NAS strategies, notably for multi-objective optimization (MOO). It signifies a paradigm shift towards a balanced pursuit of performance and energy efficiency, echoing the call for sustainable computing.

\begin{figure*}[t]
\centering
\includegraphics[width=0.33\textwidth]{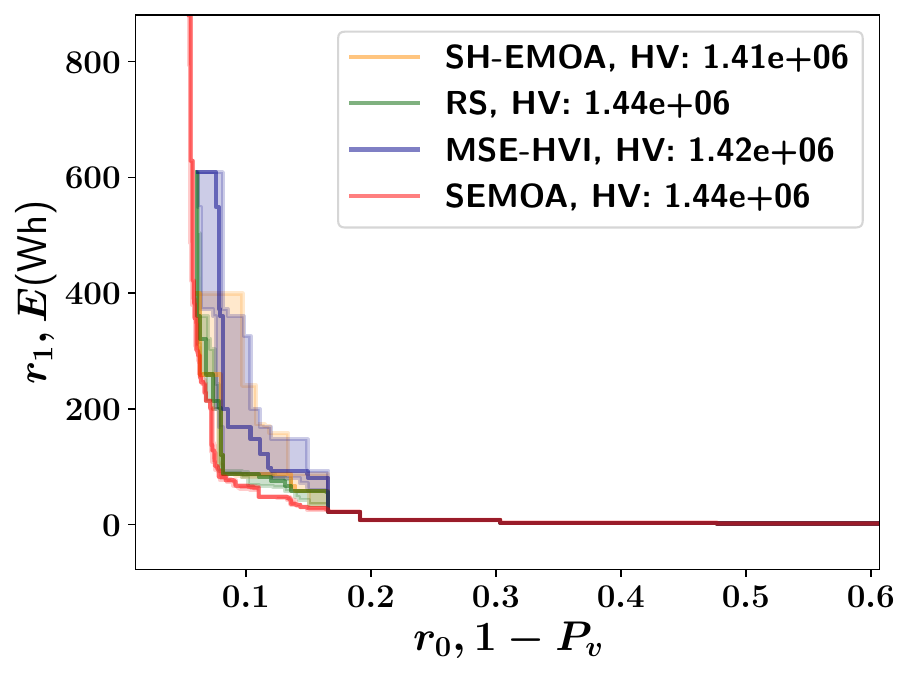}
\includegraphics[width=0.33\textwidth]{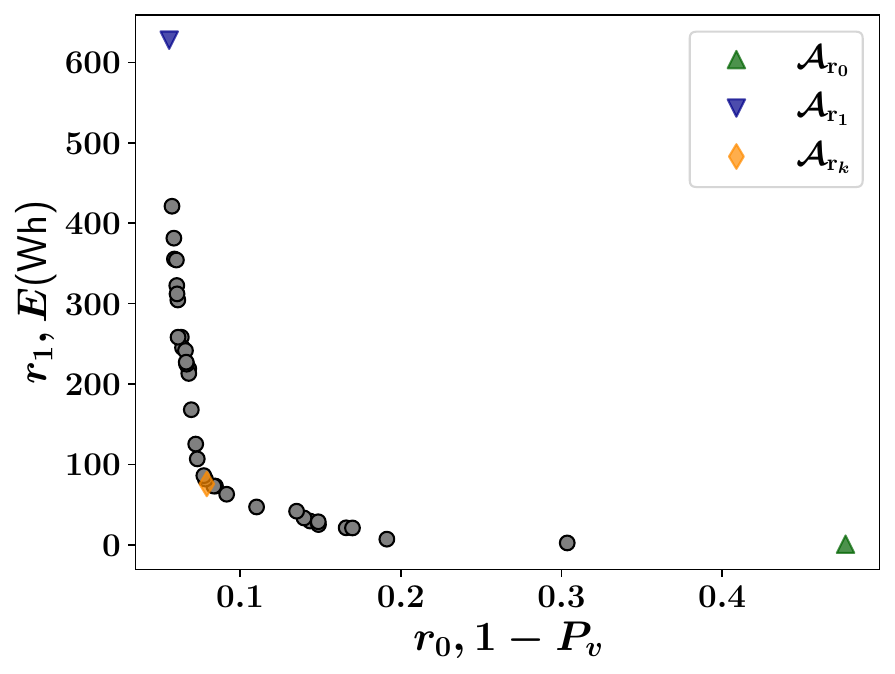}
\includegraphics[width=0.33\textwidth]{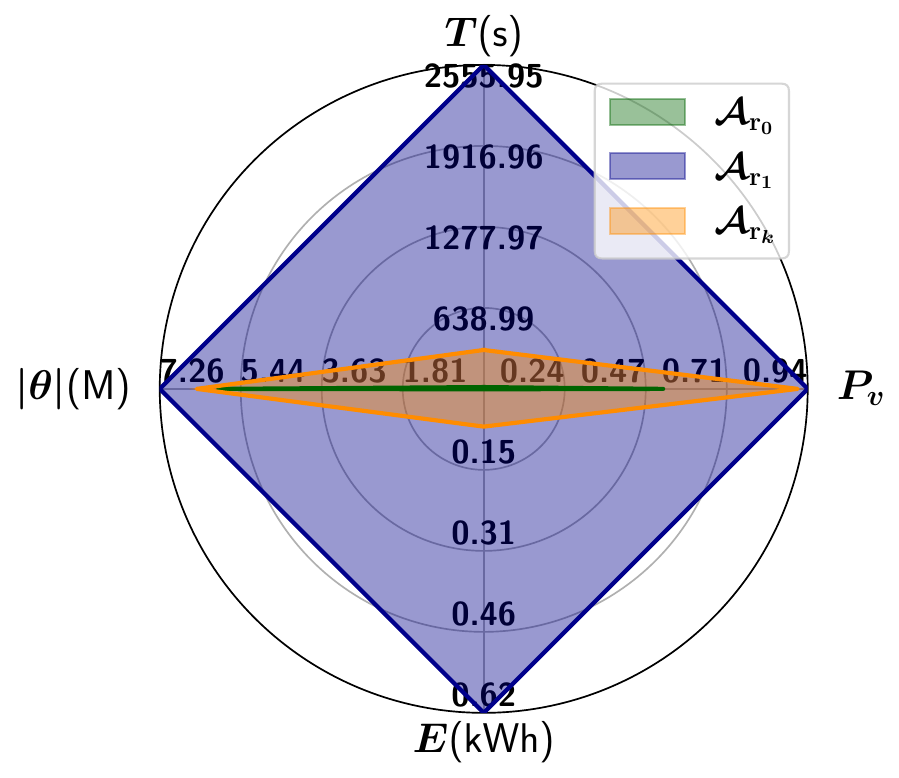}
\vspace{-0.75cm}
\caption{(Left) The attainment curve showing median solutions for 10 random initializations on the surrogate 7V space from \ecnas{} dataset. 
(Center) A representation of the Pareto front for one run of SEMOA. 
(Right) Summary of metrics for the extrema and knee point architectures for one SEMOA run.}
\label{fig:attain}
\vspace{-0.65cm}
\end{figure*}

\vspace{-0.25cm}
\section{Leveraging \ecnas{} in NAS Strategies} \label{sec:strategies}
\vspace{-0.25cm}
Tabular benchmarks like \ecnas{} offer insights into energy consumption alongside traditional performance measures, facilitating the exploration of energy-efficient architectures using multi-objective optimization (MOO) whilst emphasizing the rising need for sustainable computing.
\\
% \vspace{-0.25cm}
% \subsection{
{\bf Role of Multi-objective Optimization in NAS}: 
In the context of NAS, MOO has emerged as an instrumental approach for handling potentially conflicting objectives. We utilize the \ecnas{} benchmark to apply diverse MOO algorithms, encompassing our own simple evolutionary MOO algorithm (SEMOA) based on \cite{krause:16c} and other prominent algorithms such as SH-EMOA and MS-EHVI from \cite{izquierdo2021bag}. These methodologies are assessed against the conventional random rearch (RS) technique.

Our exploration within \ecnas{} span both single-objective optimization (SOO) and MOO. We execute algorithms across various training epoch budgets $e \in \{4,12,36,108\}$ over $100$ evolutions with a population size of $10$. For SOO, $1000$ evolutions were designated to equate the discovery potential. Results, averaged over $10$ trials, followed the methodology of \cite{izquierdo2021bag}.

For MOO, validation accuracy $P_v$ and the training energy cost, $E$ (in kWh), were chosen as the dual objectives and, for SOO, simply the performance metric. Given its indifference to parallel computing, energy consumption was chosen over training time. Inverse objectives were used for optimization in maximization tasks (e.g., $1-P_v$).
\\
% \subsection
{\bf Trade-offs in Energy Efficiency and Performance}: 
\label{sec:results}
Balancing energy efficiency with performance presents a layered challenge in NAS. Figure~\ref{fig:attain} elucidates the architectural intricacies and the prowess of various MOO algorithms in identifying energy-conservative neural architectures.

Figure~\ref{fig:attain} (left) evaluates the architecture discovery efficacy of MOO algorithms, presenting the median solutions achieved over multiple runs. SEMOA, in particular, showcases an even distribution of models attributed to its ability to exploit model locality. In contrast, SH-EMOA and MSE-HVI display a more substantial variation, highlighting the robust search space exploration of SEMOA.

The Pareto front, as depicted in Figure~\ref{fig:attain} (center), highlights extrema ($r_0, r_1$) and the knee point ($r_k$), which represents an optimal trade-off between objectives. The extrema prioritize energy efficiency or validation accuracy, while the knee point achieves a balanced feature distribution. MOO algorithms' capability to navigate the NAS space effectively is evident in their identification of architectures that balance competing objectives.

\begin{table}
\tiny
\centering
\begin{tabular}{llrclcc}
\toprule
{\bf Method} & {\bf Arch.} & $\bm{T(s)\downarrow}$ & $\bm{P_v\uparrow}$ & {\bf $\bm{E}$(kWh)}$\bm{\downarrow}$ & {\bf $\bm{|\theta|}$(M)$\bm{\downarrow}$} \\
\toprule
 & $\mathcal{A}_{\mathbf{r}_0}$ & 15.22 $\pm 2.98$ & 0.52 $\pm 0.02$ & 0.01 $\pm 0.00$ & 5.98 $\pm 0.10$ \\
{\bf SH-EMOA} & $\mathcal{A}_{\mathbf{r}_1}$ & 1034.35 $\pm 358.15$ & 0.91 $\pm 0.03$ & 0.27 $\pm 0.11$ & 6.55 $\pm 0.44$ \\
& $\mathcal{A}_{\mathbf{r}_k}$ & 226.28 $\pm 114.03$ & 0.85 $\pm 0.04$ & 0.04 $\pm 0.03$ & 6.27 $\pm 0.43$ \\
\midrule
& $\mathcal{A}_{\mathbf{r}_0}$ & 14.23 $\pm 0.00$ & 0.52 $\pm 0.00$ & 0.01 $\pm 0.00$ & 5.95 $\pm 0.00$ \\
{\bf RS} & $\mathcal{A}_{\mathbf{r}_1}$ & 1649.11 $\pm 342.02$ & 0.94 $\pm 0.00$ & 0.41 $\pm 0.09$ & 7.05 $\pm 0.18$ \\
& $\mathcal{A}_{\mathbf{r}_k}$ & 310.93 $\pm 56.50$ & 0.89 $\pm 0.03$ & 0.07 $\pm 0.03$ & 6.51 $\pm 0.38$ \\
\midrule
 & $\mathcal{A}_{\mathbf{r}_0}$ & 14.23 $\pm 0.00$ & 0.52 $\pm 0.00$ & 0.01 $\pm 0.00$ & 5.95 $\pm 0.00$ \\
{\bf MSE-HVI} & $\mathcal{A}_{\mathbf{r}_1}$ & 1112.13 $\pm 642.49$ & 0.92 $\pm 0.02$ & 0.25 $\pm 0.14$ & 6.80 $\pm 0.33$ \\
& $\mathcal{A}_{\mathbf{r}_k}$ & 191.09 $\pm 93.13$ & 0.83 $\pm 0.03$ & 0.02 $\pm 0.02$ & 6.01 $\pm 0.18$ \\
\midrule
 & $\mathcal{A}_{\mathbf{r}_0}$ & 14.23 $\pm 0.00$ & 0.52 $\pm 0.00$ & 0.01 $\pm 0.00$ & 5.95 $\pm 0.00$ \\
{\bf SEMOA}& $\mathcal{A}_{\mathbf{r}_1}$ & 2555.95 $\pm 202.42$ & 0.94 $\pm 0.00$ & 0.62 $\pm 0.08$ & 7.26 $\pm 0.15$ \\
& $\mathcal{A}_{\mathbf{r}_k}$ & 306.9 $\pm 41.86$ & 0.92 $\pm 0.01$ & 0.07 $\pm 0.01$ & 6.43 $\pm 0.06$ \\
\bottomrule
\end{tabular}
\vspace{-0.25cm}
\caption{{Average performance and resource consumption for models. Architectures $\mathcal{A}_{\mathbf{r}_0}$, $\mathcal{A}_{\mathbf{r}_1}$, and $\mathcal{A}_{\mathbf{r}_k}$ correspond to the two extrema and the knee point, respectively.}}
\label{tab:7v_metrics}
\vspace{-0.50cm}
\end{table}

\vspace{-0.25cm}
\section{Discussions}
\vspace{-0.25cm}
% \subsection
{\bf Single versus Multi-objective Optimisation}: 
Figure~\ref{fig:attain} and Table~\ref{tab:7v_metrics} capture the performance trends of solutions, elucidating that knee point solutions, $A_{r_k}$, offer architectures with about $70\%$ less energy consumption with only a $1\%$ performance degradation. Depending on specific applications, this might be an acceptable trade-off. If performance degradation is unacceptable, the Pareto front also provides alternative candidate solutions. For instance, extremum solution  $A_{r_0}$  achieves nearly the same performance as the SOO solution but consumes about $32\%$ less energy. This trend is consistent across various solutions.
\\
% \subsection
{\bf Training Time vs. Energy Consumption}:
While the original NAS-Bench-101 dataset reports training time, it cannot replace energy consumption as a metric. Even though training time generally correlates with energy consumption in single hardware regimes, the scenario changes with large-scale parallelism on multiple GPUs. Aggregate energy consumption encompasses parallel hardware and its associated overheads. Even in single GPU scenarios, energy consumption provides insights into energy-efficient models. For instance, a small architecture might consume more energy on a large GPU due to under-utilization.
\\
% \subsection
{\bf Energy-Efficient Tabular NAS Benchmarks}:
Despite the immense one-time cost of generating tabular benchmark, these benchmarks have proven highly useful for efficient evaluation of NAS strategies. For instance, our \ecnas{} dataset, predicting metrics after training models for only $4$ epochs, results in a $97\%$ reduction compared to a dataset creation from scratch. Other techniques such as predictive modelling based on learning curves \cite{yan2021bench}, gradient approximations \cite{xu2021knas}, and surrogate models fitted to architecture subsets \cite{zela2022surrogate} also prove very useful in creating new architecture spaces to consider. However, incorporating energy consumption metrics is frequently overlooked and challenges arise in integrating with existing NAS strategies. This is considering NAS benchmarks and strategies and closely intertwined, which often restricts benchmarks to tailored strategies.
% \subsection
\\
{\bf Carbon-footprint Aware NAS}:
The \ecnas{} dataset provides various metrics for each architecture. By using MOO, NAS can directly optimize the carbon footprint of models. Although instantaneous energy consumption and carbon footprint are linearly correlated, fluctuations in instantaneous regional carbon intensities can introduce discrepancies during extended training periods \cite{anthony2020carbontracker}. By reporting the carbon footprint of model training in \ecnas{}, we facilitate carbon-footprint-aware NAS \cite{selvan2022carbon}. In this work, our focus remains on energy consumption awareness, sidestepping the temporal and spatial variations of carbon intensity.
\\
% \subsection
{\bf Energy Consumption Aware Few-shot NAS}:
While tabular benchmarks like NAS-Bench-101 \cite{nasBench101Dataset} facilitate efficient exploration of various NAS strategies, they are constrained to specific architectures and datasets. Addressing this challenge involves one- or few-shot learning methods \cite{zhao2021few,nasbench1shot1}. A bridge between few-shot and surrogate tabular benchmarks emerges by combining surrogate models for predicting learning dynamics \cite{zela2022surrogate} with energy measurements. We have illustrated integrating surrogate models with existing tabular benchmarks, seamlessly extending these to surrogate benchmarks.
\\
{\bf Limitations}: Our proposed approach has certain limitations. To manage the search space, which expands exponentially with the number of vertices in the network specification DAGs, we have limited the vertices count to $\le 7$, in line with NAS-Bench-101\cite{nasBench101Dataset}. Moreover, in \ecnas{}, we utilized surrogate time and energy measurements, sidestepping the variability of training time. Despite these limitations, which primarily aim to conserve energy in experiments, the insights from these experiments can be extrapolated to broader architectural spaces.

\vspace{-0.25cm}
\section{Conclusion}
\vspace{-0.25cm}
We have enriched an established NAS benchmark by incorporating energy consumption and carbon footprint measures. \ecnas{}, encompassing $>1.6M$ entries, was crafted using an accurate surrogate model that predicts energy consumption. By showcasing Pareto-optimal solutions through MOO methods, we illuminate the potential for achieving significant energy reductions with minimal performance compromises. With its diverse metrics, \ecnas{} invites further research into developing energy-efficient and environmentally sustainable models.

%\vfill
%\pagebreak

%\vspace{-0.25cm}
% \balance
\bibliographystyle{IEEEbib}
\bibliography{strings,refs}

\input{appendix}
\end{document}

%% file: appendix.tex
\appendix
\balance

\vspace{-0.25cm}
\section{Additional Benchmarks and Metrics}
\vspace{-0.25cm}
For all benchmarks in \ecnas, we report on operations, parameter count and performance metrics, similar to NAS-Bench-101, with the addition of energy consumption. However, we introduce separate benchmarks for models characterized by DAGs with $|V| \le 4$ and $|V| \le 5$, denoted the 4V and 5V space, respectively, where we also detail the carbon footprint. For the 4V and 5V spaces the energy efficiency metrics are derived from direct measurements, independent of surrogate modeling. These datasets were compiled by performing exhaustive model training limited to 4 epochs, with the resource costs for the remaining epochs extrapolated through linear scaling. 

The primary focus for efficiency metrics is quantifying resource costs specific to model training, however, we also report total resource costs, including computational overheads, e.g., data movements. Lastly, we include the average energy consumption of computing hardware. A complete overview of the metrics relevant to this work is presented in \autoref{tab:metrics}.

\begin{table}[h]
\small
\centering
\begin{tabular}{ccc}
\toprule
\textbf{Metrics} & \textbf{Unit of measurement} & \textbf{Notation} \\
\midrule
Model parameters & Million (M) & $|\theta|$ \\
Test/Train/Eval. time & Seconds (s) & $T(s)$ \\
Test/Train/Val. Acc. & $\mathbb{R} \in [0;1] $ & $P_{v}$ \\
Energy consumption & Kilowatt-hour (kWh) & $E$(kWh) \\
Power consumption & Joule (J), Watt (W) & $E$(J), $E$(W) \\
Carbon footprint & kg\ch{CO2}eq & -- \\ 
Carbon intensity & g/kWh & -- \\
\bottomrule
\end{tabular}
\caption{Metrics reported in \texttt{EC-NAS-Bench}.}
\label{tab:metrics}
\end{table}

\section{Measurements from Carbontracker}
\vspace{-0.25cm}
\label{app:carbontracker}
Our measurements account for the energy consumption of Graphics Processing Units (GPUs), Central Processing Units (CPUs), and Dynamic Random Access Memory (DRAM), with the CPU energy usage inclusive of DRAM power consumption. Energy usage data is collected and logged at 10-second intervals, and this information is averaged over the duration of model training. The total energy consumed is then calculated and reported in kilowatt-hours (kWh), where $1 \text{kWh} = 3.6 \cdot 10^6 \text{Joules (J)}$. In addition, we assess the emission of greenhouse gases (GHG) in terms of carbon dioxide equivalents (\ch{CO2}eq), calculated by applying the carbon intensity metric, which denotes the \ch{CO2}eq emitted per kWh of electricity generated. This carbon intensity data is updated every 15 minutes during model training from a designated provider.

\begin{table}[h]
\begin{center}
\footnotesize
\begin{sc}
\begin{tabular}{cccc}
\toprule
\textbf{Space} & \textbf{Red. GPU days} & \textbf{Red. kWh} & \textbf{Red. kg\ch{CO2}eq} \\ \midrule
4V & \entoure[green]{3.758} & \entoure[green]{48.931} & \entoure[green]{6.327} \\
5V & \entoure[green]{121.109} & \entoure[green]{1970.495} & \entoure[green]{252.571} \\
7V & \entoure[green]{14037.058} & \entoure[green]{259840.907} & -- \\
\bottomrule
\end{tabular}
\end{sc}
\end{center}
\vspace{-0.5cm}
\caption{Estimated reduction in actual resource costs when creating \ecnas \ dataset for the 4V and 5V using linear scaling and 7V space using the surrogate model.}
\label{tab:reduction}
\vspace{-0.25cm}
\end{table}

However, considering only the direct energy consumption of these components does not fully capture the carbon footprint of model training, as it overlooks the energy consumption of auxiliary infrastructure, such as data centers. To address this, we refine our estimations of energy usage and carbon footprint by incorporating the 2020 global average Power Usage Effectiveness (PUE) of data centers, which stands at 1.59, as reported in \cite{pue2020}.

\begin{figure*}[t]
    \centering
    \begin{minipage}[b]{0.25\textwidth}
        \centering
        \includegraphics[width=1\textwidth]{images/radar_7v_semoa}
        SEMOA
    \end{minipage}%
    \begin{minipage}[b]{0.25\textwidth}
        \centering
        \includegraphics[width=1\textwidth]{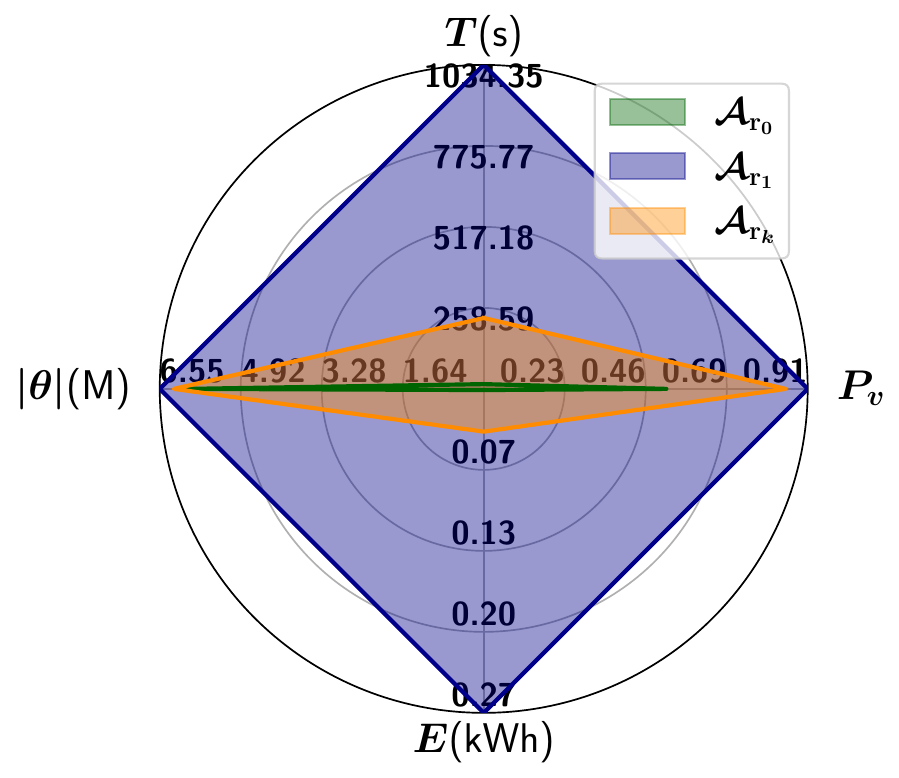}
        SH-EMOA
    \end{minipage}%
    \begin{minipage}[b]{0.25\textwidth}
        \centering
        \includegraphics[width=1\textwidth]{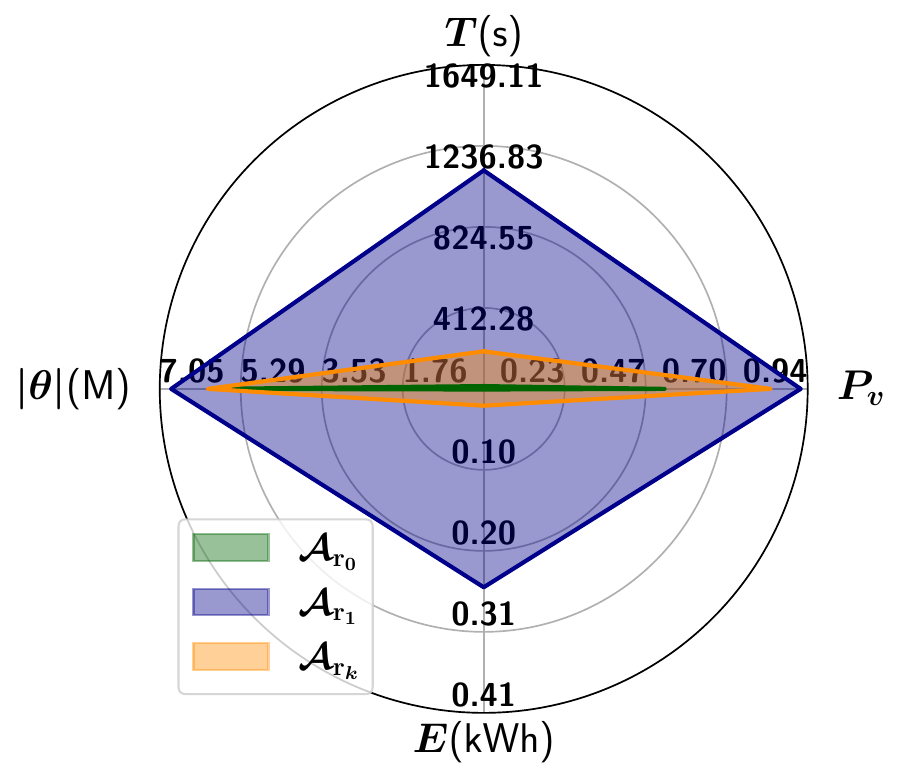}
        MSE-HVI
    \end{minipage}%
    \begin{minipage}[b]{0.25\textwidth}
        \centering
        \includegraphics[width=1\textwidth]{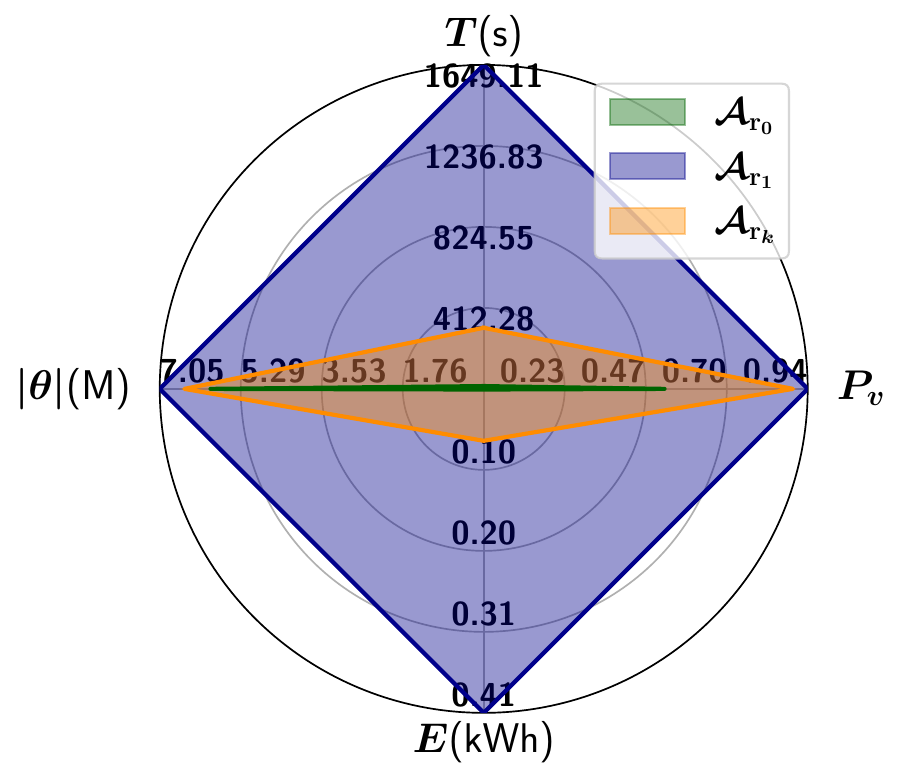}
        RS
    \end{minipage}
    \vspace{-0.65cm}
    \caption{Average performance and resource consumption across models for all baseline methods, including SEMOA. Architectures $\mathcal{A}_{\mathbf{r}_0}$, $\mathcal{A}_{\mathbf{r}_1}$, and $\mathcal{A}_{\mathbf{r}_k}$ denote the two extremes and the knee point, respectively. For precise numerical data, refer to \autoref{tab:7v_metrics}. }
    \label{fig:all_radar}
    \vspace{-0.25cm}
\end{figure*}

\section{Surrogate Model Implementation}
\vspace{-0.25cm}
\label{sec:surr_imp}
We use a simple four-layered MLP with {\tt gelu}$(\cdot)$ activation functions, except for the final layer, which transforms the input in this sequence $ 36 \rightarrow 128 \rightarrow 64 \rightarrow 32 \rightarrow  1$.

The surrogate energy model is trained using actual energy measurements from $4310$ randomly sampled architectures from the 7V space. The model was implemented in Pytorch~\cite{pytorch} and trained on a single NVIDIA RTX 3060 GPU. We use a training, validation and test split of ratio $[0.7,0.1,0.2]$ resulting in $[3020,430,860]$ data points, respectively. The MLP, $f_\theta(\cdot)$, is trained for 200 epochs with an initial learning rate of $5\times 10^{-3}$ to minimise the $L1$-norm loss function between the predicted and actual energy measurements using the Adam optimiser~\cite{kingma2015adam}.

\section{Additional Discussion}
\vspace{-0.25cm}
{\bf Resource constraind NAS.} Resource-constrained NAS for obtaining efficient architectures has been explored mainly by optimising the run-time or the number of floating point operations (FPOs). For instance, the now widely popular EfficientNet architecture was discovered using a constraint of FPOs~\cite{tan2019efficientnet}. Optimising for FPOs, however, is not entirely indicative of the efficiency of models~\cite{henderson2020towards}. It has been reported that models with fewer FPOs could have bottleneck operations that consume the bulk of the training time~\cite{howard2017mobilenets}, and some models with higher FPOs might have lower inference time~\cite{jeon2018constructing}. Energy consumption optimised hyperparameter selection outside of NAS settings for large language models has been recently investigated in~\cite{puvis-de-chavannes-etal-2021-hyperparameter}. 

\textbf{Surrogate model adaptability.} \label{app:surrmodel} Our surrogate energy model is promising in predicting energy consumption within our current search space. We have also adapted the surrogate model to the OFA search space, achieving comparable results in terms of energy consumption prediction. This suggests the potential for the surrogate model to be generalized and applied to other search spaces, broadening its applicability and usefulness in future research. Estimates for reduction in compute costs for the EC-NAS benchmark datasets are presented in \autoref{tab:reduction}.

While a comprehensive investigation of the surrogate model's performance in different search spaces is beyond the scope of this work, it is worth noting that the model could potentially serve as a valuable tool for researchers seeking to optimize energy consumption and other efficiency metrics across various architectural search spaces. Further studies focusing on the adaptability and performance of surrogate models in diverse search spaces will undoubtedly contribute to developing more efficient and environmentally sustainable AI models.

{\bf Hardware accelerators.}
Hardware accelerators have become increasingly efficient and widely adopted for edge computing and similar applications. These specialized devices offer significant performance improvements and energy efficiency, allowing faster processing and lower power consumption than traditional computing platforms. However, deriving general development principles and design directions from these accelerators can be challenging due to their highly specialized nature. Moreover, measuring energy efficiency on such devices tends to be hardware-specific, with results that may need to be more easily transferable or applicable to other platforms. Despite these challenges, we acknowledge the importance and necessity of using hardware accelerators for specific applications and recognize the value of development further to improve energy efficiency and performance on these specialized devices.

\section{Multi-objective optimisation}
\vspace{-0.25cm}
Formally, let the MOO problem be described by
$\vec{f}\colon \X \to \mathbb{R}^m,\ \vec{f}( x ) \mapsto ( f_1(x), \dots, f_m(x))$.
Here $\X$ denotes the search space of the optimisation problem and
$m$ refers to the number of objectives.  We assume w.l.o.g. that all objectives are to be minimized. For two points $x,x'\in \X$ we say that  $x'$~\emph{dominates} $x$  and write $x' \prec x$ if $\forall i \in \{ 1, \dots, m \}\colon f_i( x' ) \leq f_i( x)  \land \exists j \in \{ 1, \dots, m \}\colon  f_j( x') < f_j( x )$. For $\A,\B\subseteq \X$ we  say that $\A$ dominates $\B$ and  write $\A\prec \B$ if  $\forall x''\in \B:\exists x'\in \A: x'\prec \B$. The subset of non-dominated solutions in a set $\A\subseteq \X$ is given by $\ndom(\A)=\{x\,|\,\ x\in \A \wedge \nexists x'\in \A\setminus \{x\}:   x' \prec x \}$. The \emph{Pareto front} of a set $\A\subset\X$ defined as $\F(\A)=\{\vec{f}(x)\,|\,x\in \ndom(\A)\}$ and, thus, the goal of MOO can be formalised as approximating  $\F(\X)$.

In iterative MOO, the strategy is to step-wise improve a set of candidate solutions towards a sufficiently good approximation of $\F(\X)$. For the design of a MOO algorithm, it is important to have a way to rank two sets $\A$ and $\B$ w.r.t.\ the overall MOO goal even if neither $\A\prec \B$ nor $\B\prec \A$. This ranking can be done by the hypervolume measure. The hypervolume measure or $\mathcal S$\nobreakdash-metric (see \cite{zitzler1999multiobjective}) of a set $\A\subseteq \X$  is  the volume of the union of regions in $\mathbb{R}^m$ that are dominated by~$\A$ and bounded by some appropriately chosen reference point $\vec{r}\in \mathbb R^m$:
\[
 \displaystyle
 \HYP_{\vec{r}}( \A )
 := \VOL \left( \bigcup_{ x \in \A} \big[ f_1( x ), r_1 \big] \times
 \dots\times
 \big[ f_m( x ), r_m \big] \right),
%$
\]
%}
%\noindent
where $\VOL(\thinspace\cdot\thinspace)$ is the Lebesgue measure.
The hypervolume is, up to weighting objectives, the only strictly Pareto
compliant measure \cite{Zitzler2003} in the sense that given two sets $\A$ and~$\B$ we have $\HYP(\A) > \HYP(\B)$ if $\A$ dominates $\B$. As stated by \cite{bringmann:13}, the worst-case approximation factor of a Pareto front $\F(\A)$ obtained from any hypervolume-optimal set $\A$ with size $|\A|=\mu$ 
is asymptotically equal to the best worst-case approximation factor achievable by any set of size $\mu$, namely $\Theta(1/\mu)$ for additive approximation
and $1+\Theta(1/\mu)$ for relative approximation \cite{approxjournal}. Now we define the \emph{contributing hypervolume} of an individual $x\in \A$ as
\[
\CON_{\vec{r}}(x,\A):=
 \HYP_{\vec{r}}(\A) - 
 \HYP_{\vec{r}}(\A\setminus\{x\})\enspace.
\]%\end{equation}
The value $\CON(x,\A)$ quantifies how much a candidate solution $x$ contributed to the total hypervolume of $\A$ and can be regarded as a measure of the relevance of the point. Therefore, the contributing hypervolume is a popular criterion
in MOO algorithms \cite{beume2007sms,igel:06b,bader2011hype,krause:16c}. If we iteratively optimize some solution set $P$, then points $x$ with low $\CON(x,P)$ are candidates in an already crowded region of the current Pareto front $\F(P)$, while points with high $\CON(x,P)$ mark areas that are promising to explore further.

\subsection{SEMOA: Simple Evolutionary Multi-objective Optimisation Algorithm}
In this study, we used a simple MOO algorithm based on hypervolume maximisation outlined in Algorithm~\ref{alg:moo} inspired by \cite{krause:16c}. The algorithm iteratively updates a set $P$ of candidate solutions, starting from a set of random network architectures. Dominated solutions are removed from $P$. Then $\lambda$ new architectures are generated by first selecting $\lambda$ architectures from $P$ and then modifying these architectures according to the perturbation described in Procedure~\ref{alg:perturb}. The $\lambda$ new architectures are added to $P$ and the next iteration starts. In Procedure~\ref{alg:perturb}, the probability $p_{\text{edge}}$ for changing (i.e., either adding or removing) an edge is chosen such that in expectation, two edges are changed, and the probability $p_{\text{node}}$ for changing a node is set such that in expectation every second perturbation changes the label of a node. 

The selection of the $\lambda>m$ architectures from the current solution set 
is described in Procedure~\ref{alg:linrank}. We always select the \emph{extreme points} in $P$ that minimize a single objective (thus, the precise choice of the reference point $\vec{r}$ is of lesser importance). 
% a_1 := \operatorname*{argmin}_{a \in A}\big(\CON(a,A)\big) % := \HYP(M) - \HYP\left(M\setminus\{x\}\right),
The other $m-\lambda$ points are randomly chosen preferring points with higher contributing hypervolume. The points in $P$ are ranked according to their hypervolume contribution. The probability of being selected depends linearly on the rank. We use \emph{linear ranking selection} \cite{baker1985adaptive,greffenstette1989genetic}, where the parameter controlling the slope is set to $\eta^+=2$. Always selecting the extreme points and focusing on points with large contributing hypervolume leads to a wide spread of non-dominated solutions.

\begin{algorithm}[H]
\footnotesize
\caption{SEMOA for NAS strategy} \label{alg:moo}
\begin{algorithmic}[1]
\Require objective $\vec{f}=(f_1, \dots, f_m)$, maximum number of iterations $n$
\Ensure set of non-dominated solutions $P$
\Statex
\State Initialize $P\subset \X$ (e.g., randomly) \Comment{Initial random architectures}
\State $P \gets \ndom(P)$ \Comment{Discard dominated solutions}
\For{$i \gets 1$ to  $n$} \Comment{Loop over iterations}
\State ${O} \gets$ LinearRankSample($P$, $\lambda$) \Comment{Get $\lambda$ points from  $P$}
\State ${O} \gets$ Perturb($O$) \Comment{Change the architectures}
\State Compute $\vec{f}(x)$ for all $x\in O$\Comment{Evaluate architectures}
\State $P\gets \ndom(P \cup O)$ \Comment{Discard dominated points}
\EndFor
\State \Return $P$
\end{algorithmic}
\end{algorithm}

\floatname{algorithm}{Procedure}
% \subsection{Algorithms}
% \label{app:alg}
\begin{algorithm}
\footnotesize
\caption{Perturb($O$)} \label{alg:perturb}
\begin{algorithmic}[1]
\Require set of architectures $O$, variation probabilities for edges and nodes $p_{\text{edge}}$  and $p_{\text{node}}$ 
\Ensure set of modified architecture $O^*$
\Statex
\ForAll{$M_{\mathcal{A}} \in O$} \Comment{Loop over matrices}
\Repeat
\ForAll {$\alpha_{i, j} \in M_{\mathcal{A}}$} \Comment{Loop over entries}
\State With probability $p_{\text{edge}}$ flip $\alpha_{i, j}$
\EndFor
\ForAll {$l \in L_{\mathcal{A}}$} \Comment{Loop over labels}
\State With probability $p_{\text{node}}$ change the label of $l$
\EndFor
\Until $M_{\mathcal{A}}$ has changed
\EndFor
\State \Return $O^*$
\end{algorithmic}
\end{algorithm}

\begin{algorithm}
\footnotesize
\caption{LinearRankSample($P$, $\lambda$)} \label{alg:linrank}
\begin{algorithmic}[1]
%\Procedure{Recursion}{$a$}
\Require set $P\subset \X$ of candidate solutions, number $\lambda$ of elements to be selected; reference point $\vec{r}\in\mathbb{R}^m$, parameter controlling the preference for better ranked points $\eta^+\in [1,2]$
\Ensure $O\subset P$, $|O|=\lambda$
\Statex
\State $O=\emptyset$
\For{$i \gets 1$ to  $m$} 
\State $O\gets O \cup \operatorname{argmin}_{x\in P} f_i(x)$\Comment{Always add extremes}
\EndFor
\State Compute $\CON_{\vec{r}}(x, P)$ for all $x\in P$\Comment{Compute contributing hypervolume}
\State Sort $P$ according to $\CON(x, P)$ 
\State Define discrete probability distribution $\pi$ over $P$  where 
\[\pi_i = \frac{1}{|P|} \left(\eta^+ - 2(\eta^+-1)\frac{i-1}{|P|-1}   \right)\]
is the probability of the element $x_i$ with the $i$th largest contributing hypervolume
\For{$i \gets 1$ to  $\lambda-m$} \Comment{Randomly select remaining points}
\State Draw $x\sim \pi$\Comment{Select points with larger $\CON_{\vec{r}}$ with higher probability}
\State $O\gets O \cup x$
\EndFor
\State \Return $O$
%\EndProcedure
\end{algorithmic}
\end{algorithm}

\subsection{Multi-objective Optimization Baselines}
\label{app:moo}
{\bf Hyperparameters for the MOO Baseline Methods}
All baseline methods employ \ecnas \ for exploring and optimizing architectures. We select hyperparameters for each method to prevent unfair advantages due to increased computation time, such as the number of iterations or function evaluations. Despite allocating similar resources to the baseline methods, assessing fairness in their comparison is challenging due to the disparity in their algorithmic approaches. To mitigate uncertainties in the results, we average the outcomes over 10 experiments using different initial seeds, providing a measure of variability.

We adopt the bag-of-baselines implementation presented in \cite{izquierdo2021bag} for compatibility with the tabular benchmarks of \ecnas. Additionally, we implement the previously presented MOO algorithm SEMOA within the same framework as the baseline methods to ensure consistency. Here, we provide further details on the modifications and characteristics of the baseline methods. Summary of metrics for each method over all runs can be seen in \autoref{fig:all_radar}.

\textbf{Random Search} Unlike other methods, Random Search does not utilize evolutionary search heuristics to optimize architectures in the search space. It does not inherently consider multiple objectives but relies on processing each randomly queried model. Specifically, all queried architectures are stored, and a Pareto front is computed over all models to obtain the MOO interpretation of this method. We allow 1,000 queries for this search scheme.

\textbf{Speeding up Evolutionary Multi-Objective Algorithm (SH-EMOA)} We initialize SH-EMOA with a population size of 10 and limit the search to 100 function evaluations for budgets between 4 and 108. The algorithm is constrained to use budgets of 4, 12, 36, and 108, available in our search space. The remaining hyperparameters are set to default values, including a uniform mutation type for architecture perturbation and a tournament-style parent selection for offspring generation.

\textbf{Mixed Surrogate Expected Hypervolume Improvement (MS-EHVI)} This evolutionary algorithm is also initialized with a population size of 10 and limited to 100 evolutions. We provide an auxiliary function to discretize parameters to accommodate the experimental setup using tabular benchmarks. MS-EHVI integrates surrogate models to estimate objective values and employs the expected hypervolume improvement criterion to guide the search. This combination allows for an efficient exploration and exploitation of the search space, especially when dealing with high-dimensional and multi-objective problems.